\documentclass[10pt,twocolumn,letterpaper]{article}

\usepackage[pagenumbers]{cvpr} 

\usepackage{graphicx}
\usepackage{amsmath}
\usepackage{amssymb}
\usepackage{booktabs}
\usepackage{pbox}
\usepackage{epstopdf}
\usepackage{xspace}
\usepackage{comment}
\usepackage{lipsum}
\usepackage{bm}
\usepackage{bbm}
\usepackage{tabularx}
\usepackage{setspace}
\usepackage{sidecap}
\usepackage{xcolor}
\usepackage{wrapfig}
\usepackage{multirow}
\usepackage{array}
\usepackage{blindtext}
\usepackage{enumitem}
\usepackage{graphicx}
\usepackage{amsmath}
\usepackage{amssymb}
\usepackage{booktabs}
\usepackage{multirow}
\usepackage{diagbox}
\usepackage{hhline}
\usepackage{verbatim}
\setenumerate[1]{itemsep=0pt,partopsep=0pt,parsep=\parskip,topsep=5pt}
\setitemize[1]{itemsep=0pt,partopsep=0pt,parsep=\parskip,topsep=5pt}
\setdescription{itemsep=0pt,partopsep=0pt,parsep=\parskip,topsep=5pt}
\usepackage[pagebackref,breaklinks,colorlinks]{hyperref}

\usepackage[capitalize]{cleveref}
\crefname{section}{Sec.}{Secs.}
\Crefname{section}{Section}{Sections}
\Crefname{table}{Table}{Tables}
\crefname{table}{Tab.}{Tabs.}

\def\cvprPaperID{4155} 
\def\confName{CVPR}
\def\confYear{2023}

\begin{document}

\title{Complete-to-Partial 4D Distillation for Self-Supervised Point Cloud Sequence Representation Learning}


\author{
Zhuoyang Zhang\thanks{Equal contribution with the order determined by rolling dice.}\\
IIIS, Tsinghua University\\
\and
Yuhao Dong\footnotemark[1]\\
Department of Automation \\ Tsinghua University\\
\and
Yunze Liu\\
IIIS, Tsinghua University\\
Shanghai Qi Zhi Institute\\
\and
Li Yi \thanks{Corresponding author.}\\
IIIS, Tsinghua University\\
Shanghai Artificial Intelligence Laboratory\\
Shanghai Qi Zhi Institute\\
}
\maketitle

\begin{abstract}
Recent work on 4D point cloud sequences has attracted a lot of attention. However, obtaining exhaustively labeled 4D datasets is often very expensive and laborious, so it is especially important to investigate how to utilize raw unlabeled data. However, most existing self-supervised point cloud representation learning methods only consider geometry from a static snapshot omitting the fact that sequential observations of dynamic scenes could reveal more comprehensive geometric details. And the video representation learning frameworks mostly model motion as image space flows, let alone being 3D-geometric-aware. To overcome such issues, this paper proposes a new 4D self-supervised pre-training method called Complete-to-Partial 4D Distillation. Our key idea is to formulate 4D self-supervised representation learning as a teacher-student knowledge distillation framework and let the student learn useful 4D representations with the guidance of the teacher. Experiments show that this approach significantly outperforms previous pre-training approaches on a wide range of 4D point cloud sequence understanding tasks including indoor and outdoor scenarios.
\end{abstract}

\section{Introduction}
\begin{figure}[t]
    \centering
    \includegraphics[width=0.85\linewidth]{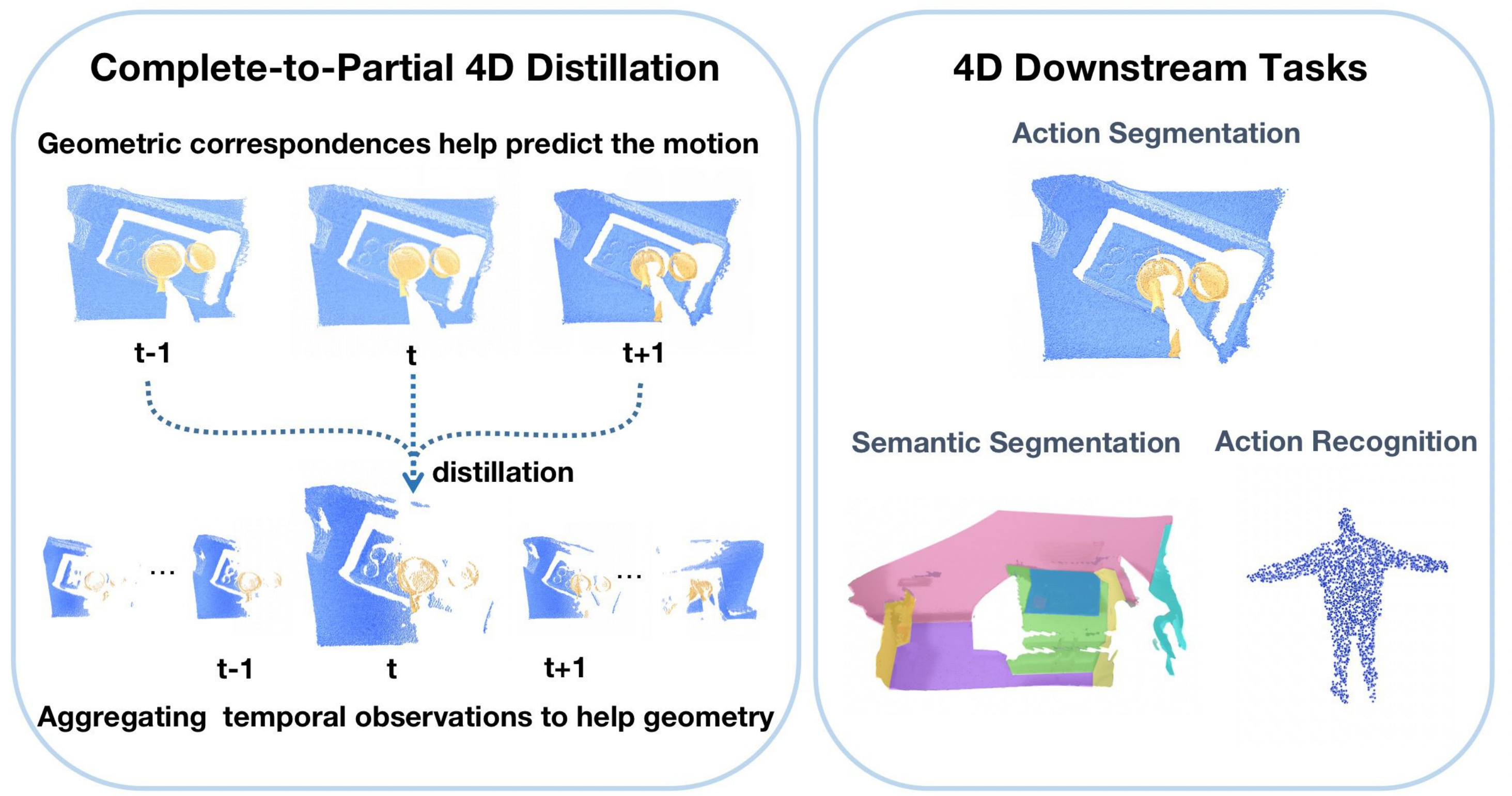}
    \caption{We propose a complete-to-partial 4D distillation (C2P) approach. Our key idea is to formulate 4D self-supervised representation learning as a teacher-student knowledge distillation framework in which students learn useful 4D representations under the guidance of a teacher. The learned features can be transferred to a range of 4D downstream tasks.}
    \label{fig:teaser}
    \vspace{-5mm}
\end{figure}
 Recently, there is a surge of interest in understanding point cloud sequences in 4D (3D space + 1D time)~\cite{hoi4d,wen2022point,p4transformer,pstnet,https://doi.org/10.48550/arxiv.2207.04673}. As the direct sensor input for a wide range of applications including robotics and augmented reality, point cloud sequences faithfully depict a dynamic environment regarding its geometric content and object movements in the context of the camera ego-motion. Though widely accessible, such 4D data is prohibitively expensive to annotate in large scale with fine details. As a result, there is a strong need for leveraging the colossal amount of unlabeled sequences. Among the possible solutions, self-supervised representation learning has shown its effectiveness in a wide range of fields including images~\cite{DBLP:journals/corr/abs-2111-06377,DBLP:journals/corr/abs-2002-05709,DBLP:journals/corr/abs-1911-05722}, videos~\cite{sermanet2018time,huang2021ascnet,https://doi.org/10.48550/arxiv.2205.09113,DBLP:journals/corr/abs-2101-07974,DBLP:journals/corr/abs-2112-03587} and point clouds~\cite{DBLP:journals/corr/abs-2109-00179, DBLP:journals/corr/abs-2007-10985,zhang2021pointclip, https://doi.org/10.48550/arxiv.2205.14401,pang2022masked}. We therefore aim to fill in the absence of self-supervised point cloud sequence representation learning in this work.

Learning 4D representations in a self-supervised manner seems to be a straightforward extension of 3D cases. However, a second thought reveals its challenging nature since such representations need to unify the geometry and motion information in a synergetic manner. From the geometry aspect, a 4D representation learner needs to understand 3D geometry in a dynamic context. However, most existing self-supervised point cloud representation learning methods~\cite{https://doi.org/10.48550/arxiv.2205.14401,DBLP:journals/corr/abs-2109-00179,zhang2021pointclip,pang2022masked} only consider geometry from a static snapshot, omitting the fact that sequential observations of dynamic scenes could reveal more comprehensive geometric details. From the motion aspect, a 4D representation learner needs to understand motion in the 3D space, which requires an accurate cross-time geometric association. Nevertheless, existing video representation learning frameworks~\cite{https://doi.org/10.48550/arxiv.2205.09113,DBLP:journals/corr/abs-2101-07974,DBLP:journals/corr/abs-2112-03587} mostly model motion as image space flows, letting alone being 3D-geometric-aware. Due to such challenges, 4D has been rarely discussed in the self-supervised representation learning literature, with only a few works~\cite{chen2022_4dcontrast,liu2020p4contrast} designing learning objectives in 4D to facilitate static 3D scene understanding.

To address the above challenges, we examine the nature of 4D dynamic point cloud sequences, and draw two main observations. First, most of a point cloud sequence depicts the same underlying 3D content with an optional dynamic motion. Motion understanding could help aggregate temporal observations to form a more comprehensive geometric description of the scene. Second, geometric correspondences across time could help estimate the relative motion between two frames. Therefore better geometric understanding should facilitate a better motion estimation. At the core are the synergetic nature of geometry and motion.

To facilitate the synergy of geometry and motion, we develop a Complete-to-Partial 4D Distillation (C2P) method. Our key idea is to formulate 4D self-supervised representation learning as a teacher-student knowledge distillation framework and let the student learn useful 4D representations with the guidance of the teacher.
And we present a unified solution to the following three questions: How to teach the student to aggregate sequential geometry for more complete geometric understanding leveraging motion cues? How to teach the student to predict motion based upon better geometric understanding? How to form a stable and high-quality teacher?

In particular, our C2P method consists of three key designs as shown in Figure~\ref{fig:teaser}. 
First, we design a partial-view 4D sequence generation method to convert an input point cloud sequence which is already captured partially into an even more partial sequence. This is achieved by conducting view projection of input frames following a generated camera trajectory. The generated partial 4D sequence allows bootstrapping multi-frame geometry completion. This is achieved by feeding the input sequence and the generated partial-view sequence to teacher and student networks respectively and distill the teacher knowledge to a 4D student network.
Second, the student network not only needs to learn completion by mimicking the corresponding frames of the teacher network, but also needs to predict the teacher features of other frames within a time window, to achieve so-called 4D distillation. Notice the teacher feature corresponds to more complete geometry, which also encourages the student to exploit the benefit of geometry completion in motion prediction.
Finally, we design an asymmetric teacher-student distillation framework for stable training and high-quality representation, i.e., the teacher network has weaker expressivity compared with the student but is easier to optimize. 

We evaluate our method on four downstream tasks including indoor and outdoor scenarios: 4D action segmentation on HOI4D~\cite{hoi4d}, 4D semantic segmentation on HOI4D~\cite{hoi4d}, 4D semantic segmentation on Synthia 4D~\cite{synthia} and 3D action recognition on MSR-action3D~\cite{msr}. We demonstrate significant improvements over the previous method($+2.5\%$ accuracy on HOI4D action segmentation, $+1.0\%$ mIoU on HOI4D semantic segmentation, $+1.0\%$ mIoU on Synthia 4D semantic segmentation and $+2.1\%$ accuracy on MSR-Action3D).

In summary, our contributions can be listed below:
\begin{itemize}
    \item We propose a new 4D self-supervised representation learning method named Complete-to-Partial 4D Distillation which facilitates the synergy of geometry and motion learning.
    \item We propose a natural and effective way to generate partial-view 4D sequences and demonstrate that it can work well as learning material for knowledge distillation.
    \item We find that asymmetric design is crucial in the complete-to-partial knowledge distillation process and we propose a new asymmetric distillation architecture. 
    \item Extensive experiments on four tasks show that our method outperforms previous state-of-the-art methods by a large margin.
\end{itemize}

\begin{figure*}[t]
    \centering
    \includegraphics[width=0.7\linewidth]{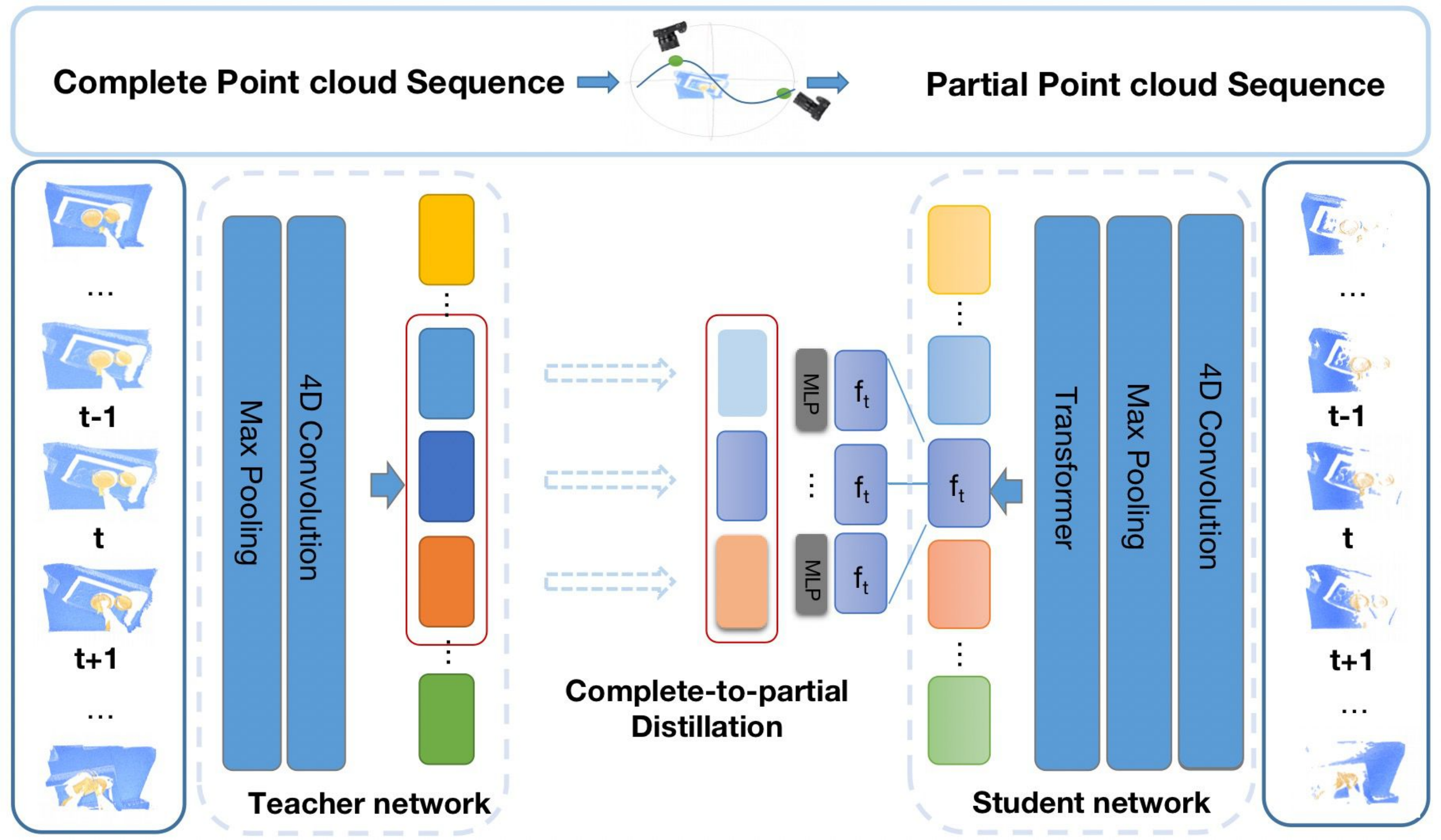}
    \caption{Overview of our method. Our method can be divided into three main parts, partial-view generation, asymmetric twin network feature extraction, and complete to partial 4D knowledge distillation. (a) In the partial-view generation part, our goal is to generate a sequence of partial-views related to the perspective. (b) Asymmetric teacher-student network can perform 4D knowledge distillation in a better and more stable way. (c) Then the knowledge from the teacher network within a time window is distilled into the features of a frame extracted by the student network.}
    \label{fig:overview}
    \vspace{-5mm}
\end{figure*}   

\section{Related Work}
\subsection{4D Point Cloud Sequence Understanding}
Unlike 3D static scenes, Understanding 4D point cloud sequences require more focus on leveraging spatiotemporal information to perceive complete geometry and sensitive motion. There are many 4D sequence-based tasks~\cite{synthia,hoi4d,msr} that have received extensive attention, such as 4D semantic segmentation~\cite{synthia, hoi4d}, 3D action recognition~\cite{msr}, 4D action segmentation~\cite{hoi4d}, etc. These tasks tend to have high computational overhead, as 4D data require a large memory occupation. According to the representation, existing 4D backbone can be categorized into voxel-based~\cite{lin2018toward,minkowski}, and point-based methods~\cite{p4transformer,pstnet,meteornet,wen2022point}. The state-of-the-art approach is based on the transformer architecture~\cite{p4transformer, wen2022point}, which is often difficult to optimize and requires a large amount of labeled data for training. An under-explored problem for 4D sequence understanding is how to learn spatio-temporal features in an unsupervised manner to reduce the difficulty of network optimization and the amount of labeled data required. We proposed a new unsupervised contrastive knowledge distillation approach to learn 4D representation. 



\subsection{3D Representation Learning}
Benefiting from the rapid development of 2D representational learning~\cite{DBLP:journals/corr/abs-1911-05722,DBLP:journals/corr/abs-2111-06377,liu2021contrastive}, 3D representational learning~\cite{pang2022masked,yu2021pointbert,zhang2021pointclip,liu2020p4contrast} has also been widely explored. Existing methods can be grouped into generative-based methods~\cite{DBLP:journals/corr/abs-2006-12373,OcCo,yu2021pointbert,https://doi.org/10.48550/arxiv.2205.14401}, context-based methods~\cite{DBLP:journals/corr/abs-2007-10985,DBLP:journals/corr/abs-2108-07794,zhang_depth_contrast}. For generative-based~\cite{DBLP:journals/corr/abs-2006-12373,OcCo,yu2021pointbert} methods, PSG-Net~\cite{DBLP:journals/corr/abs-2006-12373} propose to learn representation by reconstructing point cloud objects with seed generation. OcCo~\cite{OcCo} utilize a U-net to complete occluded point cloud objects. Point-Bert~\cite{yu2021pointbert} learn the representation for Transformers by recovering masked object parts. For context-based methods, PointContrast~\cite{DBLP:journals/corr/abs-2007-10985} propose contrasting different views of scene point clouds at point-level. CSC~\cite{hou2021exploring} adopt a spatial partition to improve pointcontrast. RandomRooms~\cite{DBLP:journals/corr/abs-2108-07794} construct pseudo scenes with synthetic objects for contrastive learning. SelfCorrection~\cite{chen2021shape} learn the informative representation by distinguishing and restoring destroyed objects. DepthContrast~\cite{zhang_depth_contrast} extend to contrast with multiple representations (points and voxels) at scene-level. Due to the high dimensionality of 4D data and the required computational overhead, most of the existing methods cannot be directly extended to 4D data. DCGLR~\cite{fu2022distillation} is most relevant to us, which first constructs the global point cloud set and the local point cloud set by cropping the full point cloud with different cropping ratios and then distilling knowledge from the local branch to the global branch. Our goal is to investigate how to extract high-quality spatio-temporal features at the scene-level to help 4D downstream tasks, so we propose to use distill 4D complete information to the partial branch using contrastive learning.
\subsection{4D Representation Learning}
4D Representation Learning is still a relatively new field, and thanks to the experience in 3D, researchers have gradually started to focus on how to exploit information from 4D sequences. 4Dcontrast~\cite{chen2022_4dcontrast} proposes to exploit 4D motion information to improve the effectiveness of 3D tasks. However, due to the memory overhead, it is difficult to use this point-level comparison learning method on long 4D sequences. STRL~\cite{huang2021spatio} uses temporal-spatial contrastive learning to learn good representations of 3D objects. Although pre-trained on 4D data, both of these methods are designed to learn static 3D representations. Our approach broadens this scope to use 4D data for pre-training to improve the understanding of dynamic scenes, and to our knowledge, our approach is also the first exploration in this direction.

\section{Method}

In this paper, we propose a new 4D self-supervised representation learning method named Complete-to-Partial 4D Distillation. To avoid ambiguity, we first emphasize the definitions of complete and partial. The "complete" represents an original input 4D point cloud sequence, which was obtained through the natural collection. A "complete point cloud" may still be a geometrically incomplete point cloud, but it can be fully described as complete compared to the new 4D point cloud sequence we generated. A "partial" represents a synthetic point cloud generated from the captured data, which is significantly less complete than the original point cloud, so we call it a "partial 4D point cloud sequence". In this paper, the definitions of partial and partial-view are not distinguished since we are generating new data through the camera view.

Our main idea is to distill the spatial-temporal information of a complete point cloud sequence into a partial point cloud sequence so that the neural network can extract strong features in a self-supervised manner. 

The overview of our method is shown in Figure \ref{fig:overview}. Our method is divided into three main parts, partial-view generation, asymmetric twin network feature extraction, and complete to partial 4D knowledge distillation. In the partial-view generation part, our goal is to generate a sequence of partial-views related to the perspective.
Since this approach can generate a wide variety of 4D sequences, various motion patterns during the point cloud change can be simulated. After this, the partial-view sequences are fed to the student network and the complete point cloud sequences are fed to the teacher network. 
We design an asymmetric teacher-student network to perform 4D knowledge distillation in a better and more stable way. The teacher network has a weaker representational capacity, which facilitates the student network to distill stable information in an efficient manner. Finally, the knowledge from the teacher network within a time window is distilled into the features of a frame extracted by the student network, which makes the student network not only learn the geometric completeness but also capture the motion information from the spatial-temporal context and the geometric change information of the foreground-background. Since the student network is distilling spatio-temporal features from a time window of the teacher network, we call the method a 4D knowledge distillation framework. 

In the rest of this section, we will introduce our method in detail. We first show how to generate the partial point cloud sequence in section~\ref{sec:generation}. Then we present the asymmetric teacher-student network design in section~\ref{sec:Asymmetric}. In section~\ref{sec:distill}, We illustrate in detail how to distill 4D spatio-temporal information from complete point cloud sequences to partial point cloud sequences. 

\begin{figure}[t]
    \centering
    \includegraphics[width=0.85\linewidth]{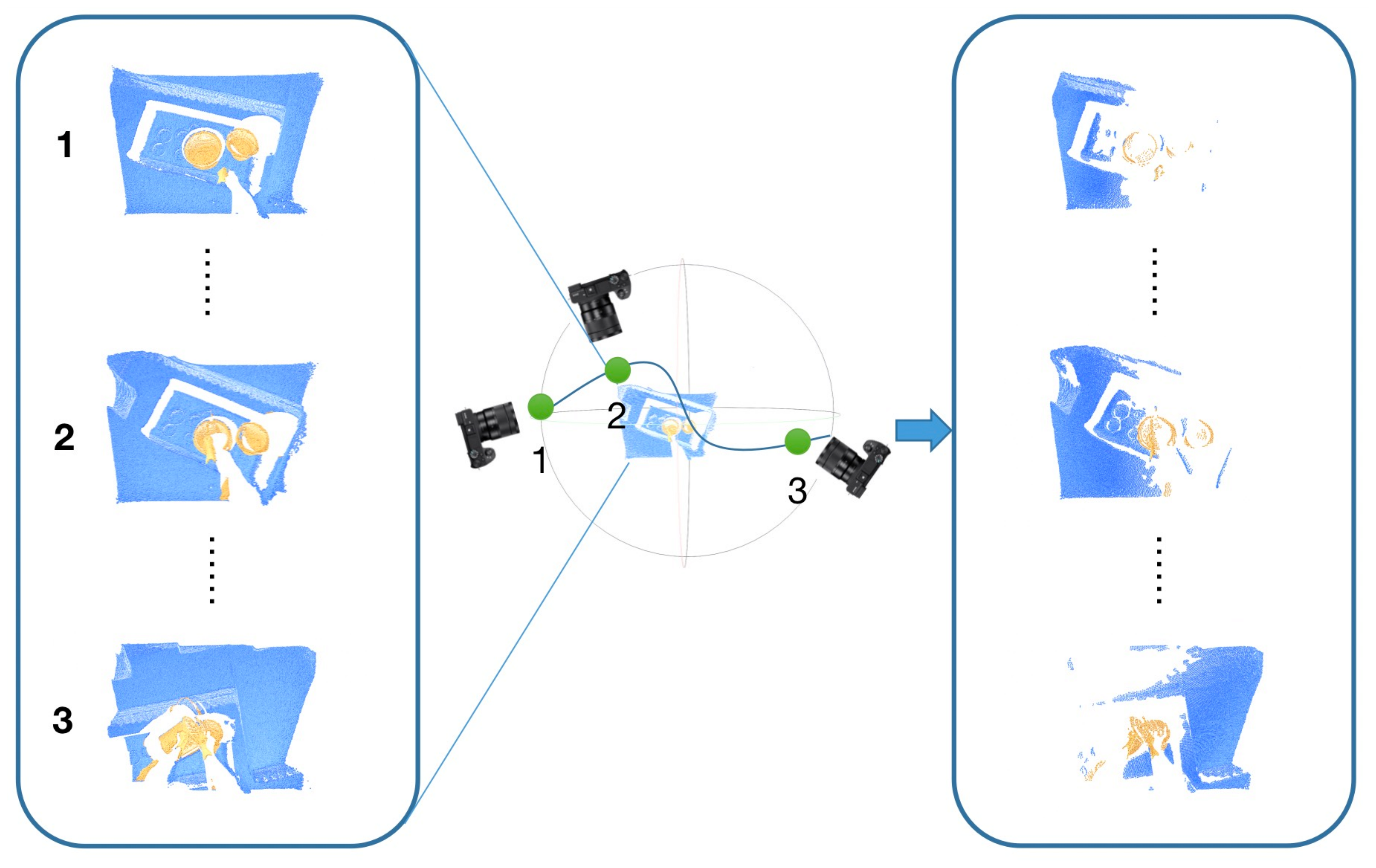}
    \caption{Illustration of partial-view sequences generation.}
    \label{fig:partial}
    \vspace{-5mm}
\end{figure}

\subsection{Generating partial-view 4D sequences}
\label{sec:generation}
A static 3D point cloud can generate new point clouds with different views. These new data can reflect the geometric properties of the same 3D object/scene in a complementary way. Inspired by this observation, different 4D sequences can be generated according to different trajectories. We think it is a natural idea to use different motion trajectories in 4D data to generate a large number of partial-view 4D sequences. These new trajectories are all descriptions of the same "complete 4D sequence", but the differences in their poses and occlusion patterns can reflect the real physical world comprehensively. 

We define the "complete point cloud sequence" with sequence length L as $S = \{s_1, s_2, \dots, s_L\}$ where each $s_i$ denotes one frame. For each frame, it consists of N points, $P_i = \{p^i_{1}, p^i_{2}, \dots, p^i_{N}\}$, where each point $p^i$ is a vector of both coordinates and other features such as color and normal. We generate the "partial point cloud sequence" by sampling a natural camera trajectory and occluding invisible points from the camera view. Figure~\ref{fig:partial} show the process of partial-view 4D sequences generation. And we will describe how to sample natural camera trajectory and how to do occlusion sampling in detail.

\textbf{Sampling Natural Camera Trajectory.}
\label{sec:snct}
Based on the camera trajectory we can generate a new 4D sequence and sample the visible points according to the camera pose of each frame, which naturally fits the real physical world better than random sampling. Moreover, random sampling is not consistent, which hinders the network from learning spatio-temporal contextual information from the data. We believe that a natural camera trajectory can simulate the motion of the 3D world well, so the new 4D data generated is more conducive to learning spatio-temporal features of the network. We also verified this in our experiments. 

Specifically, we first determine a sphere on which our trajectory can be approximated as a curve. The center of the sphere is the same as the center of the point cloud and the radius is the distance between the center of the point cloud and the original camera position. For each position on the sphere, it can actually be determined by two angles $\theta$ and $\phi$. The $\theta$ determines the angle on the horizontal plane and $\phi$ determines the angle on the vertical plane. We define the sphere coordinates of the original camera position as $(\frac{\pi}{2}, 0)$. For the horizontal movement, we have two movement modes: one is from $-\frac{5}{12}\pi$ to $\frac{5}{12}\pi$ (left to right from the original camera view), and the other is from $\frac{5}{12}\pi$ to $-\frac{5}{12}\pi$ (right to left from the original camera view). Each move pattern has a horizontal angle change of 150 degrees, and we set the angle change evenly for adjacent frames. For the vertical movement, we assign only $\pm$ 5 degrees of continuous interference to the trajectory. In addition to horizontal and vertical movement, randomly zooming in and out is also included in our trajectory.

\textbf{Occlusion Sampling.}
Based on each camera track, we can sample visible points and drop invisible points based on occlusion relationships. This generates a more realistic 4D partial-view point cloud. Occlusion sampling is done per frame with the same camera intrinsic and different camera view-point. Given a camera with its view-point and intrinsic matrix, we need to first transform the point cloud in the world coordinate system into the camera coordinate system. suppose the intrinsic matrix is K, and the view-point is described by [{R}$\mid${t}] where ${R}$ is the rotation matrix and {t} is the translation vector. Then for a point $ P = (x, y, z)^T $ in the world coordinate system, the position $\Tilde{P} = (\Tilde{x}, \Tilde{y}, \Tilde{z})^T$ in the camera coordinate system will be $\Tilde{P}$ = {K}[{R}$\mid${t}][P] where [P] is the screw representation $(x, y, z, 1)^T$. After the projection, we obtain a new 4D sequence that is realistic and natural. 
In the camera coordinate system, we will project the point clouds back into a depth image with 2D pixels. Invisible points will have the same 2D pixel coordinate as visible points but have a larger depth value. In this way we can determine the points to be occluded and get the partial-view depth image. Then we just map the depth image back into point clouds via the inverse view change transformation and get the partial point cloud.
Repeating the above operation, we can easily obtain a large number of camera traces and the corresponding 4D partial-view sequences.

\subsection{Asymmetric Distillation Architecture}
\label{sec:Asymmetric}
Most traditional knowledge distillation frameworks use twin networks to act as teacher-student networks. Since most of the current 4D backbone of state of the art is based on Transformer's architecture, influenced by its training stability, we experimentally found that it is difficult for the student network to learn stable and meaningful features when using a symmetric twin network. A teacher network with weaker representational abilities facilitates a better distillation of knowledge for the student network. Unlike previous 2D/3D work, we designed an asymmetric knowledge distillation framework to perform stable and efficient training. Specifically, we design the student network as a point 4D Conv layer followed by a transformer layer, while the teacher network contains only a point 4D Conv layer.
The transformer is used for the student network to leverage motion information to find the geometric correlation, which can help construct complete geometry. While for the teacher branch, a temporal transformer is not needed because the target of the student learning is a complete geometry and motion-sensitive feature, so a local feature extracting for the teacher branch is already enough. Letting the teacher branch contain the temporal transformer will mess up the local feature representation and make the feature difficult for students to learn. Our experiments observed this observation and verified the reasonableness of the design.

\subsection{Complete to Partial 4D Distillation}
\label{sec:distill}
This section focuses on how to teach student networks to use motion cues to aggregate continuous geometric information for a more complete geometric understanding, and how to teach student networks to predict motion based on a better geometric understanding. 

\begin{figure}[t]
    \centering
    \includegraphics[width=0.85\linewidth]{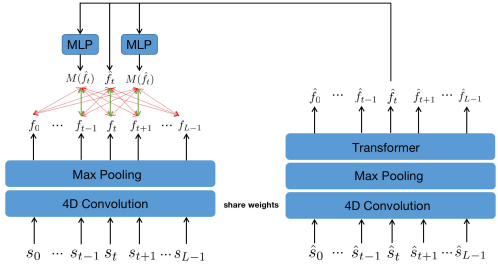}
    \caption{Objective of our proposed contrastive distillation. }
    \label{fig:objective}
    \vspace{-5mm}
\end{figure}

Based on this, we design a 4D-to-4D distillation framework as shown in Figure \ref{fig:objective}. 
On the one hand, our feature extraction backbone is a 4D backbone, and our goal is to teach it to efficiently and stably utilize spatial and temporal information in order to recover the underlying 3D real world. On the other hand, 4D means that the distillation source is not only a single-frame feature, but a window of features across time. From the single frame, we can only distill static geometric information, even if the backbone network is 4D. Whereas from a temporal window, the network can better reconstruct the geometry of the current frame from temporal cues, and can predict motion based on the geometric consistency of adjacent frames.

\textbf{Frame-level Feature Extraction.} 
The overall above process is executed at a frame-level feature.
For nowadays state-of-the-art transformer-based 4D backbones, a point cloud will be divided into several tokens and feature communicates at the token-level. 

We believe frame-level features can be learned better in 4D networks, so we do the complete to partial distillation on the frame-level features instead of at the token or point level. This is due to the irregular representation and high down-sample ratio of the point cloud sequence. 
For the token level, it is very difficult to align tokens in different frames precisely under nowadays popular token-generating methods(use furthest point sampling to first generate anchor points and then ball-query to search near-neighbor points to extract token features). 
The error of alignment may seriously destroy the learning process. For the point level, a high down-sample ratio will make the point cloud even more sparse and irregular. The same region may get a totally different point cloud pattern after high ratio downsampling so it is very hard for the network to predict raw points directly. We also verified this experimentally.
The frame-level information exchange allows for better learning of global geometric integrity and motion, which is critical for 4D downstream tasks.

Formally, given a complete point cloud sequence $S = \{s_1, s_2, ..., s_L\}$ and a partial point cloud sequence $\hat{S} = \{\hat{s}_1, \hat{s}_2, ..., \hat{s}_L\}$, the teacher network and student network will take these two sequences as input separately and get two sequences of per-frame features $F = \{f_1, f_2, ..., f_L\}$, $\hat{F} = \{\hat{f}_1, \hat{f}_2, ..., \hat{f}_L\}$. Then for the i-th frame of the partial sequence where the student feature is $\hat{f}_i$, its distillation source is $D = \{\dots,f_{i-1}, f_{i}, f_{i+1},\dots\}$. And we denote the dynamic predictors as $M$. 

For geometry, the i-th frame features extracted by the student network need to be learned directly from the i-th frame information of the teacher network, which is a process of learning the complete point cloud from the partial-view point cloud. Since the features are extracted by a 4D backbone, this process actually encourages the network to reconstruct the complete geometry based on temporal cues. For time, the i-th frame extracted by the student network needs to have the ability to predict information to non-corresponding frames within a time window of the teacher network. Taking a time window of 3 as an example, we use two prediction heads to predict the i-1 frame and the i+1 frame of the teacher network from the i-th frame, respectively. This process encourages the network to learn the geometric correlation between adjacent frames through knowledge distillation of motional information.

\textbf{4D Contrastive Learning Supervision.} Since contrast learning shows excellent performance on the organized feature space, we use contrastive learning to supervise the feature learning process. 
For geometry, the i-th frame feature drawn by the student network is used as the anchor sample, and its positive sample is the i-th frame feature drawn by the teacher network. The other frames outside the time window form the negative sample pool. We assume our time window is $D = \{\dots,f_{i-1}, f_{i}, f_{i+1},\dots\}$. The set of frame indexes within the time window we denote by ${W_t}$.
\begin{equation}
    {L}_{geo} = -\sum_{i} log\frac{\text{exp}(\hat{f}_i\cdot f_i/\tau )}{\sum_{j\notin W_t } \text{exp}(\hat{f}_i\cdot f_j/\tau )}
\end{equation}
For time, the ith frame of the student network is used as the anchor, and the frame features within the time window are first obtained after the prediction head. As an example, the positive samples of frame t-1 and t+1 obtained from the predictor are the frame t-1 and t+1 features from the teacher's network, respectively, and the same for the other frames. The negative samples are the pool of features from all frames outside the time window.
\begin{equation}
    {L}_{time} = -\sum_{i} log\frac{\text{exp}(M(\hat{f}_i)\cdot  f_i/\tau )}{\sum_{j\notin W_t } \text{exp}(M(\hat{f}_i)\cdot f_j/\tau )}\\
\end{equation}
where $\tau$ is the temperature coefficient.

So the final loss function is:
\begin{equation}
    {L}_{total} = \alpha_1{L}_{geo}+\alpha_2{L}_{time}
\end{equation}
where $\alpha_1$, $\alpha_2$ are the coefficient that $\alpha_1 + \alpha_2 = 1$.

By constructing the above objective, We encourage the student network to learn geometric completeness and temporal motility from the teacher network simultaneously.

\section{Experiments}

In this section, following other representation learning methods, we use the pre-trained network weights from each task as initialization and fine-tune them in 4D downstream tasks. The performance gain will be a good indicator of measuring the quality of the learned feature. In this section, we cover four 4D point cloud sequence understanding tasks: action segmentation on HOI4D~\cite{hoi4d}, semantic segmentation on HOI4D~\cite{hoi4d}, semantic segmentation on Synthia 4D~\cite{synthia} and 3D action recognition on MSR-action3D~\cite{msr} in Section~\ref{sec:HOI4Das}, \ref{sec:HOI4Dss}, \ref{sec:synthia} and \ref{sec:recog} respectively. For these four tasks, some basic settings will first be introduced in Section \ref{sec:cbs}. In addition, we provide extensive ablation studies to validate our design choices in Section~\ref{sec:ablation}.

\subsection{Basic Setting}
\label{sec:cbs}
\textbf{Partial-view Sequence Generating.} To guarantee sufficient partial-view observations, we always make the camera go around the point cloud with a 150 degrees horizontal angle($\theta$) change. The horizontal angle change between two adjacent frames is the same. We set randomly the moving direction so it has the equal possibility for the camera to move from $\theta = -\frac{5}{12}\pi$ to $\frac{5}{12}\pi$ or $\theta = \frac{5}{12}\pi$ to $-\frac{5}{12}\pi$. Vertical angle($\phi$) change and zoom-in/zoom-out disturbance setting follow the strategy we introduce in Section \ref{sec:snct}. Note that due to the different point cloud properties in Synthia 4D dataset, we use a slightly different generating strategy which will be further explained in Section \ref{sec:synthia}.

\textbf{Distillation Network.} We set the time window size as 3 consisting of frames i-1, i, i+1 through all the tasks. P4Transformer and PPTr are used as the backbone by default. Little differences in the backbone across different tasks will be further clarified in each section. For the dynamic predictor, we use two fully connected layers together with one layer norm layer and a rectification layer. There are two dynamic predictors for predicting the t+1 frame and t-1 frame separately. 

\begin{table}
\setlength{\tabcolsep}{0.15mm}
\scriptsize{
\begin{center}
\caption{Action segmentation on HOI4D dataset~\cite{hoi4d}}
\label{table:hoi4das}
\newcolumntype{Y}{>{\centering\arraybackslash}X}
{
\begin{tabularx}{\columnwidth}{>{\centering} m{0.3\columnwidth}|>{\centering} m{0.12\columnwidth}|>{\centering} m{0.125\columnwidth}|Y|Y|Y|Y}
\hline\noalign{\smallskip}
Method & Frames   & Acc & Edit & F1@10 & F1@25 & F1@50\\
\noalign{\smallskip}
\hline
\noalign{\smallskip}

P4Transformer~\cite{p4transformer} & 150 & 71.2 & 73.1 & 73.8 & 69.2& 58.2\\
P4Transformer+C2P~\cite{p4transformer} & 150 & 73.5 & 76.8 & 77.2 & 72.9 & 62.4\\
\noalign{\smallskip}
\hline
\noalign{\smallskip}
PPTr~\cite{wen2022point} & 150 &  77.4 & 80.1 & 81.7 & 78.5 & 69.5\\
PPTr+STRL~\cite{DBLP:journals/corr/abs-2109-00179} & 150 & 78.4 & 79.1 & 81.8 & 78.6 & 69.7 \\
PPTr+VideoMAE~\cite{https://doi.org/10.48550/arxiv.2205.09113} & 150 & 78.6 & 80.2 & 81.9 & 78.7 & 69.9\\
PPTr+C2P~\cite{wen2022point} & 150  & \textbf{81.1} & \textbf{84.0} & \textbf{85.4} & \textbf{82.5} & \textbf{74.1}\\
\noalign{\smallskip}
\hline
\end{tabularx}
}
\vspace{-6mm}
\end{center}
}
\end{table}

\subsection{Fine-tuning on HOI4D Action Segmentation}
\label{sec:HOI4Das}
\textbf{Setup.} To demonstrate the effect of our approach, we first conduct experiments on the HOI4D action segmentation task. For each point cloud sequence, we need to predict the action label for each frame. We follow the official data split of HOI4D with 2971 training scenes and 892 test scenes. Each sequence has 150 frames, and each frame has 2048 points. We use the version of PPTr without primitive fitting as the backbone, which is released by its author. However, due to the new emergence of the action segmentation task in 4D, there is no specific model for action segmentation of P4Transformer and PPTr before, we do a little change to the model of action recognition to adapt to the action segmentation task. The significant change is that per frame feature is formed before the temporal transformer, so attention is performed on frame features instead of token features as designed in the former backbone. And the output of the model will be a sequence of labels instead of one label. We have verified this design is more effective from the backbone design view. All experiments in action segmentation will use this model design. We also conduct experiments with other 4D pre-training strategies including STRL(a 4D spatial-temporal contrastive learning method) and VideoMAE(an MAE-based method for video representation learning which can be easily extended to 4D data). The following metrics are reported: framewise accuracy (Acc), segmental edit distance, as well as segmental F1 scores at the overlapping thresholds of 10\%, 25\%, and 50\%. Overlapping thresholds are determined by the IoU ratio. 

\textbf{Result.} As reported in Table~\ref{table:hoi4das}, our method has big improvement on both two backbones. For the state-of-the-art backbone PPTr, it can be seen that our method consistently outperforms STRL and VideoMAE by a big margin for all metrics. Considering STRL, its short sequence augmentation can not well guide the model to notice motion cues so it is very hard to leverage temporal information which is actually very important in action segmentation tasks. Simple extension of VideoMAE to point cloud sequence also shows very little improvement. Unlike video pixel tokens which are regular and compact, high down-sampled point cloud tokens have very irregular and sparse patterns. It is hard for the model to learn to predict raw points. Also notice that VideoMAE do self-supervised learning on a point level, which is very hard to learn motion features. Our method emphasizes motion information by cross-time distillation and the learning process is done on a stable frame feature level which finally comes in a satisfying result. 

We further give a visualization to intuitively demonstrate the outstanding
performance of our method. As shown in Figure \ref{fig:AS_vis}, each row from the top to the bottom: ground truth, train from scratch, pretrain with VideoMAE, pretrain with STRL and pretrain with C2P. Results show that our method outperforms the
others in the continuity and accuracy of the segmentation, which results in a better performance on all the metrics.

\begin{table}
\setlength{\tabcolsep}{0.15mm}
\scriptsize{
\begin{center}
\caption{Semantic segmentation on HOI4D dataset~\cite{hoi4d}}
\label{table:hoi4dss}
\newcolumntype{Y}{>{\centering\arraybackslash}X}
{
\begin{tabularx}{\columnwidth}{>{\centering} m{0.47\columnwidth}|>{\centering} m{0.2\columnwidth}|Y}
\hline\noalign{\smallskip}
Method & Frames   & mIoU\\
\noalign{\smallskip}
\hline
\noalign{\smallskip}

P4Transformer~\cite{p4transformer} & 3	 & 40.1 \\
P4Transformer+C2P~\cite{p4transformer} & 3 & 41.4 \\
\noalign{\smallskip}
\hline
\noalign{\smallskip}
PPTr~\cite{wen2022point} & 3 & 41.0 \\
PPTr+STRL~\cite{DBLP:journals/corr/abs-2109-00179} & 3  & 41.2 \\
PPTr+VideoMAE~\cite{https://doi.org/10.48550/arxiv.2205.09113} & 3 & 41.3 \\
PPTr+C2P~\cite{wen2022point} & 3  & \textbf{42.3} \\
\noalign{\smallskip}
\hline
\end{tabularx}
}
\vspace{-6mm}
\end{center}
}
\end{table}

\subsection{Fine-tuning on HOI4D Semantic Segmentation}
\label{sec:HOI4Dss}

\textbf{Setup.} To verify that our approach can also be effective on fine-grained tasks, we conducted further experiments on HOI4D for 4D semantic segmentation. Since pre-training methods generally benefit from large data, unlike previous papers, we use the full set of HOI4D for our experiments. The dataset consists of 3863 4D sequences, each including 300 frames of point clouds, for a total of 1.158M frames of point clouds. For one frame, there are 8192 points. We follow the official data split of HOI4D with 2971 training scenes and 892 test scenes. We use the version of PPTr without primitive fitting as the backbone, which is released by its author. During representation learning and training/fine-tuning, we randomly select 1/5 of the whole data to form one epoch for efficient training. We use mean IoU(mIoU) \% as the evaluation metric and 39 category labels are used to calculate it.

Considering our representation learning method prefers long sequence which has relatively abundant temporal information while the limitation of GPU memory, we set the sequence length as 10 and num points per frame as 4096. Fine-tuning and testing are performed on sequence length of 3 to be consistent with the baseline. 



\textbf{Result.} As reported in Table~\ref{table:hoi4dss}, there is still a performance improvement on the 4D semantic segmentation task, which also shows the effectiveness of our approach for fine-grained feature understanding.
The very small improvement that STRL can provide suggests that simple data augmentation at the scene-level does not help much in learning fine-grained features.
VideoMAE performs one convolution layer on pixels to form token features so it is very easy to match the token with the raw points. However, there are several convolution layers in the semantic segmentation model so it is very hard to figure out the corresponding token and the raw points which may cause serious information leakage that hurt the pretraining.
Compared with previous methods, our method can effectively extract features with high representational and generalization capabilities by introducing 4D distillation.

\subsection{Fine-tuning on Synthia 4D Semantic Segmentation}
\label{sec:synthia}

\textbf{Setup.} To further show that our method is also
effective in outdoor scenarios, we conduct experiments on Synthia 4D dataset. Different from the former experiments where point clouds are produced from depth images which are incomplete, point clouds in Synthia 4D dataset are complete. This facilitates us to move the camera around the point cloud with a 360 degrees horizontal angle($\theta$) change. Specifically, we set the pre-training sequence length as 12, so there is a 30 degrees horizontal angle change between two adjacent frames.

Synthia 4D\cite{synthia} is a synthetic dataset generated from Synthia dataset\cite{synthia}. It consists of six videos of driving scenarios where both objects and cameras are moving. Following previous work\cite{p4transformer}, we use the same training/validation/test split, with 19,888/815/1,886 frames, respectively. The pre-training clip length is 12 with 4096 points in each frame. Fine-tuning is done on clip length 3 with 16384 points in each frame to keep consistent with the previous methods. We use P4Transformer\cite{p4transformer} as the backbone. The mean Intersection over Union (mIoU) is used as the evaluation metric.

\textbf{Result.} As reported in Table~\ref{table:synthia4d}, our method has a considerable improvement. This indicates that our method is also general to outdoor scenarios. Specifically, we observe that our method has a large improvement on several small objects which further reflects that the network has a better understanding of geometry and motion.

\newcommand{\tabincell}[2]{\begin{tabular}{@{}#1@{}}#2\end{tabular}} 
\begin{table}
\setlength{\tabcolsep}{0.2mm}
\tiny{
\begin{center}
\caption{Evaluation for semantic segmentation on Synthia 4D dataset~\cite{synthia}}
\label{table:synthia4d}
\begin{tabular}{l|c|cccccccccccc|c}
\hline\noalign{\smallskip}
Method & Frames & Bldn& Road& Sdwlk & Fence& Vegittn & Pole & Car & T.Sign & Pedstrn & Bicycl & Lane & T.Light& mIoU\\
\noalign{\smallskip}
\hline
\noalign{\smallskip}
P4Transformer~\cite{p4transformer} & 1 & 96.76 & 98.23 & 92.11 & 95.23 & \textbf{98.62} & 97.77 & 95.46 & 80.75 & 85.48 & 0.00 & 74.28 & 74.22 & 82.41\\
P4Transformer~\cite{p4transformer} & 3 & 96.73 & 98.35 & \textbf{94.03} & 95.23 & 98.28 & 98.01 & 95.60 & 81.54 & 85.18 & 0.00 & 75.95 & 79.07 & 83.16 \\
\noalign{\smallskip}
\hline
\noalign{\smallskip}
\tiny{P4Transformer+C2P~\cite{p4transformer}} & 3 & \textbf{97.02} & \textbf{98.54} & 93.21 & \textbf{95.52} & 97.80 & \textbf{98.12} & \textbf{95.87} & \textbf{84.81} & \textbf{88.19} & 0.00 & \textbf{77.62} & \textbf{82.60} & \textbf{84.11} \\
\noalign{\smallskip}
\hline
\end{tabular}
\vspace{-6mm}
\end{center}
}
\end{table}

\subsection{Fine-tuning on 3D Action Recognition}
\label{sec:recog}
\textbf{Setup.} Following P4Transformer and PPTr, we used the MAR-Action3D dataset, which consists of 567 human point cloud videos, including 20 action categories. Each frame is sampled with 2,048 points. The point cloud videos are segmented into multiple segments. During training, video-level labels are used as segment-level labels. To estimate the video-level probabilities, we take the mean of all segment-level probability predictions. To be able to compare with the best performance, We fit the human point cloud to 4 primitives and then use PPTr as our 4D backbone. Our pretraining is done on the sequence of length 24, and the pre-trained model weights are used for all the other sequence lengths. 

\textbf{Result.} The results are shown in Table~\ref{table:msraction3d}. We can observe that we improve the performance by a large margin for different sequence lengths which demonstrates the effectiveness of our method. We observe that STRL and VideoMAE also don not have a significant improvement as the same in action segmentation. 
This indicates that our method significantly outperforms the existing methods in terms of the ability to extract global features of sequences.

\begin{table}
\setlength{\tabcolsep}{1mm}
\scriptsize{
\begin{center}
\caption{Action recognition on MSR-Action3D dataset~\cite{msr}}
\label{table:msraction3d}
\begin{tabular}{l|c|c|c}
\hline\noalign{\smallskip}
Method & Input & Frames & Video Acc@1\\
\noalign{\smallskip}
\hline
\noalign{\smallskip}
PointNet++~\cite{pointnet2}& point& 1 & 61.61 \\
\noalign{\smallskip}
\hline
\noalign{\smallskip}
& point& 8 & 81.14\\
MeteorNet~\cite{meteornet} & point& 16 & 88.21\\
& point& 24 & 88.50\\
\noalign{\smallskip}
\hline
\noalign{\smallskip}
& point& 8 & 83.50\\
PSTNet~\cite{pstnet} & point& 16 & 89.90\\
& point& 24 & 91.20\\
\noalign{\smallskip}
\hline
\noalign{\smallskip}
& point& 8 & 83.17\\
P4Transformer~\cite{p4transformer} & point& 16 & 89.56\\
& point& 24 & 90.94\\
\noalign{\smallskip}
\hline
\noalign{\smallskip}
& point& 8 & 84.02\\
PPTr~\cite{wen2022point} & point&  16& 90.31 \\
& point&  24& 92.33\\
\noalign{\smallskip}
\hline

\noalign{\smallskip}
PPTr+STRL~\cite{DBLP:journals/corr/abs-2109-00179} & point& 24 & 92.66 \\
PPTr+VideoMAE~\cite{https://doi.org/10.48550/arxiv.2205.09113} & point& 24 & 92.66 \\
\noalign{\smallskip}
\hline

\noalign{\smallskip}
& point& 8 & 87.16 \\
PPTr+C2P& point& 16& 91.89 \\
& point&  24& \textbf{94.76} \\
\noalign{\smallskip}
\hline

\end{tabular}

\end{center}
}
\end{table}


\subsection{Analysis Experiments and Discussions}
\label{sec:ablation}
In this section, we first conduct an ablation study to verify the design of complete to partial.
We then discuss the differences between symmetric and asymmetric distillation frameworks.
We also compare the effects of random sampling and sampling based on natural camera trajectories.
We use HOI4D Action Segmentation as the downstream task and PPTr~\cite{wen2022point} as the default backbone. For more other ablation studies, please check \ref{sec:more}.


\textbf{Necessity of complete to partial distillation.}
We propose a complete-to-partial distillation framework, where the teacher network uses complete point cloud sequences to teach the student network that has only seen partial-view sequences. This knowledge distillation not only encourages networks to use temporal cues to achieve an understanding of the complete geometry, but also to use geometric consistency to capture motion information. This is a capability that neither the complete-to-complete nor the partial-to-partial distillation frameworks possess. On the one hand, geometric complementarity between multiple frames from the same data is difficult to help the understanding of the current frame due to the lack of teacher material that can be used as input for teacher network. On the other hand, the teacher network and the student network have seen almost the same data with little point cloud variation, which greatly reduces the difficulty of the self-supervised task. All these problems hinder the integration of spatio-temporal information. To demonstrate this, we experiment with three different distillation strategies. Frame-wise accuracy achieved by the three strategies are 79.48, 79.66, 81.10 respectively as shown in Table \ref{tab:complete_to_partial}, and our Complete-to-Partial strategy achieves the best performance.   
\begin{table}[h]
\centering
\caption{Comparison of different distillation strategies.}
\label{tab:complete_to_partial}
\begin{tabular}{c|c} 
    \toprule
    Methods  & Frame-wise accuracy\\
    \midrule 
    Complete-to-Complete  & 79.48\\
    Partial-to-Partial &  79.66\\
    Complete-to-Partial & \textbf{81.10}\\
    \bottomrule
\end{tabular}
\vspace{-10pt}
\end{table}



\textbf{Symmetric versus asymmetric framework.}
One of the more popular frameworks for knowledge distillation is the use of symmetric twin networks, where the teacher network and the student network are the same and share weights. The student network needs to learn the features extracted by the teacher network. In our setup, both the teacher and student networks are 4D backbone networks based on the structure of point 4D Conv (student and teacher networks) and Transformer (student network). The data source of the teacher is a complete point cloud sequence and the data source of the student network is a partial point cloud sequence. we believe that a teacher network with weaker representational abilities facilitates the student network to better distill knowledge. And we expect that the student should be better than the teacher network, i.e., have stronger learning and representational capabilities than the teacher network. Therefore, we design the student network with an additional Transformer layer than the teacher network. To verify this, we re-trained the teacher network using a symmetric twin network, i.e., adding a Transformer layer to the teacher network.
The results show that when the teacher network is the same as the student network, at best only an accuracy of 78.95 can be obtained, which is 2.15 lower than the asymmetric structure. the performance degradation is probably due to the instability of the teacher network during the training process.
The above experiments also demonstrate the importance of asymmetric network design in complete to partial distillation.

\textbf{Partial-view generation strategy.}
We design a partial-view sequence generation method that simulates camera motion in the natural world, i.e., first generating camera trajectories and then sampling visible points according to occlusion relations. We believe that this approach can help the student network generate effective learning input. With this natural sequence, the student network can make full use of the temporal information to understand the complete geometric, and can use the learned geometry to understand the motion of the object/scene.
The learning inputs we generate are significantly more meaningful and effective than those obtained by random sampling. The random sampling approach sacrifices the geometric consistency of the point cloud and the reality of the data, both of which have a negative impact on knowledge distillation. 
We conducted experiments to verify this. Using random sampling to generate 4D point cloud sequences, the same framework can only achieve an accuracy of 80.22. This verifies that our partial-view generation approach is reasonable and effective.









\section{Conclusion}
In this paper, we propose complete-to-partial 4D distillation, a new pre-training method for point cloud sequence representation learning. 
Our main idea is to distill the spatial-temporal information of a complete point cloud sequence into a partial point cloud sequenes. Experiments show our proposed method significantly outperforms previous pre-training approaches on a wide range of point cloud sequence understanding tasks. 
Although the pre-training of point cloud sequences is still at an early stage, this problem is undoubtedly very important and challenging.
Our work is encouraging and suggests future work to explore more possible design for 4D representation learning.


{\small
\bibliographystyle{main}
\bibliography{main}
}

\appendix
\setcounter{section}{0}
\def\thesection{\Alph{section}}

\section{Data-efficient 4D Representation Learning}
\label{sec:data-eff}

\begin{figure}[h]
    \centering
    \includegraphics[width=1.0\linewidth]{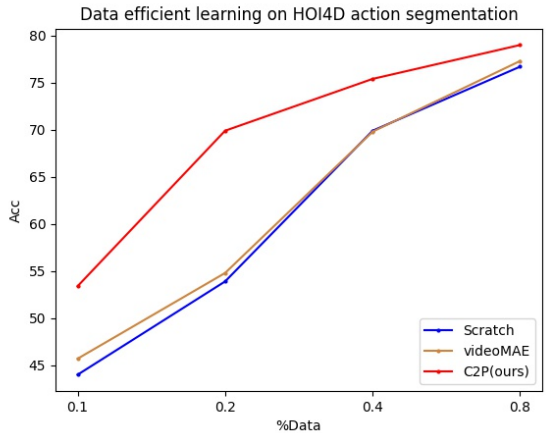}
    \caption{Data-efficient learning on HOI4D Action Segmentation. Our C2P method turns out to be more competitive as the number of data decreases, while training from scratch and the VideoMAE method shows significant performance degradation. }
    \label{fig:data_efficient}
    \vspace{-5mm}
\end{figure}

We evaluate our method under limited training data on the HOI4D Action Segmentation task. For all data-efficient experiments, our limited data are randomly sampled from the full dataset of HOI4D Action Segmentation dataset. For pre-training and fine-tuning, we use the same setup as we describe in Section 4 in the main paper.  As shown in Table \ref{table:hoi4das_efficient} and Figure \ref{fig:data_efficient}. Our pre-training method shows consistently outstanding performance in the case of lack of data, compared with VideoMAE\cite{https://doi.org/10.48550/arxiv.2205.09113}. This indicates that our method can still formulate a good 4D representation under limited data.

\begin{table}[h]
\setlength{\tabcolsep}{0.15mm}
\scriptsize{
\begin{center}
\caption{Data-efficient learning on HOI4D Action Segmentation.}
\label{table:hoi4das_efficient}
\newcolumntype{Y}{>{\centering\arraybackslash}X}
{
\begin{tabularx}
{\columnwidth}{>{\centering} m{0.15\columnwidth}|>{\centering} m{0.2\columnwidth}|>{\centering} m{0.3\columnwidth}|Y}
\hline\noalign{\smallskip}
\%Data & Scratch   & VideoMAE\cite{https://doi.org/10.48550/arxiv.2205.09113} & C2P(ours)\\
\noalign{\smallskip}
\hline
\noalign{\smallskip}

10\% & 44.0 & $45.7_{\textcolor[rgb]{0.5,0.5,0.5}{(+1.7)}}$ & $53.4_{\textcolor[RGB]{18,220,168}{(+9.4)}}$ \\
20\% & 53.9 & $54.8_{\textcolor[rgb]{0.5,0.5,0.5}{(+0.9)}}$ & $69.9_{\textcolor[RGB]{18,220,168}{(+16.0)}}$ \\
40\% & 69.9 & $69.8_{\textcolor[rgb]{0.5,0.5,0.5}{(-0.1)}}$ & $75.4_{\textcolor[RGB]{18,220,168}{(+5.5)}}$\\
80\% & 76.7 & $77.3_{\textcolor[rgb]{0.5,0.5,0.5}{(+0.6)}}$ & $79.0_{\textcolor[RGB]{18,220,168}{(+2.3)}}$ \\
\noalign{\smallskip}
\hline
\end{tabularx}
}
\end{center}
}
\end{table}

\section{Other Analysis Experiments}
\label{sec:more}
\subsection{Different Time Windows}
\label{sec:time-window}
We believe a sequence-to-sequence distillation framework encourages the network not only to perceive motion information through geometric consistency but also to obtain complete geometric understanding on the basis of temporal cues. So the time window needs to be carefully designed because it determines the source of knowledge for sequence-to-sequence distillation. We set the time windows to 1 (3D distillation), 3 (our setting), and 5 for experiments on action segmentation. 

Results are shown in Table \ref{tab:time_window}. When the size of the time window is set to 1, 3D distillation is clearly not enough to learn a good 4D representation due to the sacrifice of the benefits from temporal information. 
When the time window size is set to 3, the network is able to learn temporal information between multiple frames, resulting in a better ability to leverage temporal information and better performance. 
As we continue to increase the time window to 5, we observe some performance degradation which may resulted by the complexity and difficulty of optimization. Specifically, per-frame prediction needs more mlp-heads, which makes the network not easy to integrate spatial-temporal information among different frames.To verify this, we conduct another experiment with time window size set to 5. We use two predictors to predict the t-2 frame and t+2 frame respectively. As shown in Table \ref{tab:time_window}, we get a result similar to the best performance. The above experiments show that our method likewise gains improvement with a larger time window size, but the problem of optimization should be taken into account in deciding which frames are selected to do the distillation. 

\begin{table}[h]
\centering
\caption{Comparison of different time window size.}
\label{tab:time_window}
\begin{tabular}{c|c} 
    \toprule
    Window Size  & Accuracy\\
    \midrule 
    1  & 79.84\\
    3 &  \textbf{81.10}\\
    5 & 79.76\\
    5 ($\pm$2 frames) & 80.75 \\
    \bottomrule
\end{tabular}
\end{table}

\subsection{A Single MLP or Multiple MLPs}
\label{sec:mlp}
In our 4D-to-4D distillation framework, the network learns to aggregate temporal information by predicting frame-wise features. Considering the difference between frames, we use frame-wise predictors to gain better prediction results. We believe frame-wise predictor plays a positive role in attaining better spatial-temporal information. Experiments have been conducted to verify the impact of different predictor choices. For a time window size set to 3, we use a single mlp-head predictor to replace the frame-wise predictor and get a degradation of 0.92 than the original result, indicating frame-wise predictors are more capable of capturing spatial-temporal information. With the time window size set to 5, We use two predictors to predict the previous two frames and the latter two frames respectively. We get a degradation of 0.66 compared with the results shown in Table \ref{tab:time_window}. Experiments show that single predictor for multiple frames makes it difficult to aggregate cross-time information and our frame-wise predictor is quite important for a better integration of spatial-temporal information.  
\begin{table}[h]
\centering
\caption{Comparison of mlp-head choice.}
\label{tab:mlp_choice}
\begin{tabular}{c|c} 
    \toprule
    Window Size  & Accuracy\\
    \midrule 
    1 mlp for 3 frames  & 80.18\\
    2 mlps for 5 frames &  80.09\\
    \bottomrule
\end{tabular}
\end{table}




\section{Implementation Details}
\label{sec:implementation}
In this section, we introduce the details of the implementation of HOI4D Action Segmentation experiments.

\textbf{Pre-training setup.}
We use SGD optimizer to train the network. The learning rate is set to be 0.01 and we use a learning rate warmup for 10 epochs, where the learning rate increases linearly for the first 10 epochs. The dimension of the features used to calculate the contrastive loss is set to 2048. The temperature $\tau$ used when calculating contrastive loss is set to 0.07. The time window size is set to 3. For the combination of the loss, the loss from the current frame accounts for fifty per cent and the loss from the previous frame and the latter accounts for twenty-five per cent respectively. As for P4DConv, we set the spatial stride to 32, the radius of the ball query is set to 0.9 and the number of samples is set to 32 by default. With batch-size set to 8, our pre-training method can be implemented on two NVIDIA GeForce 3090 GPUs.

\textbf{Fine-tuning setup.}
We use SGD optimizer to fine-tune the network. The learning rate is set to be 0.05 and we use a learning rate warmup for 5 epochs likewise. To achieve better performance, we apply learning rate decay on 20 and 35 epochs with a decay ratio set to 0.5. For model parameters, the hyper-parameters of the model are exactly the same as pre-training. With batch-size set to 8, our fine-tuning method can be implemented on two NVIDIA GeForce 3090 GPUs.

\begin{figure*}[t]
    \centering
    \includegraphics[width=1.0\linewidth]{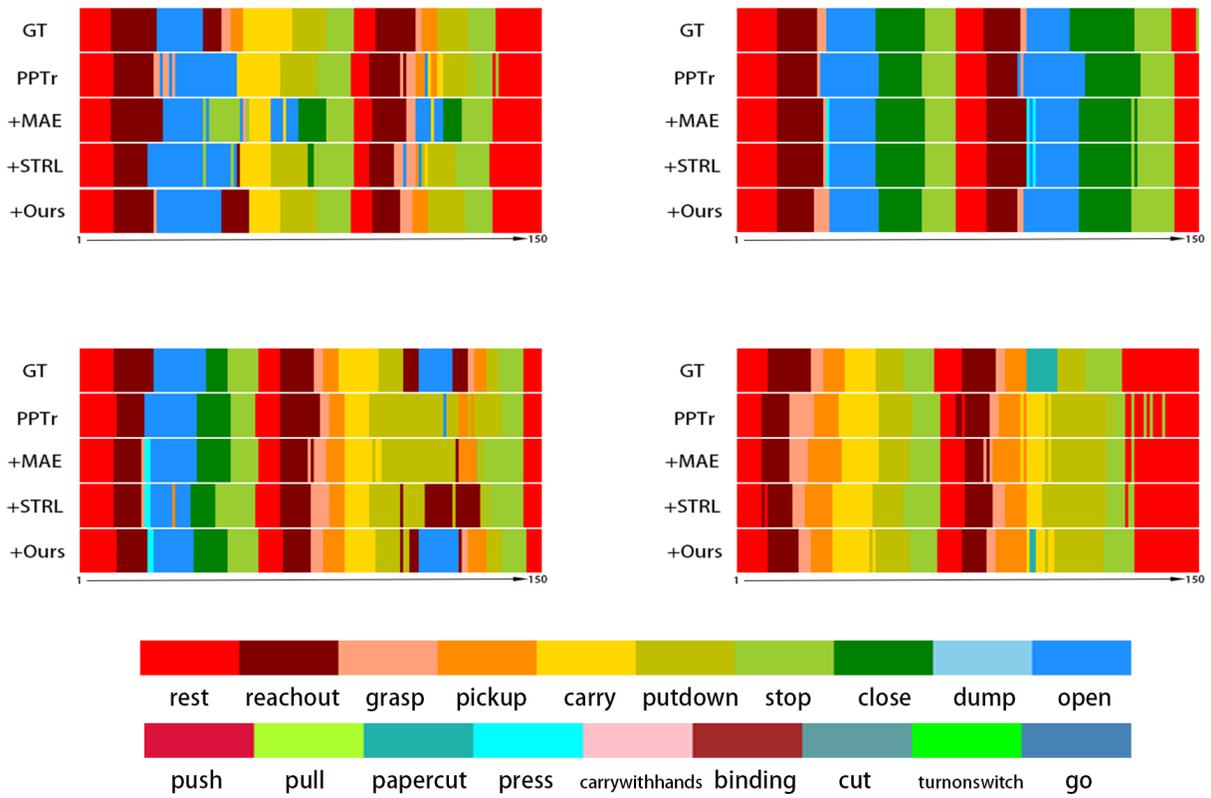}
    \caption{Visualization of results of action segmentation on HOI4D.}
    \label{fig:AS_vis}
\end{figure*}

\end{document}